\documentclass[11pt]{article}

\usepackage{amsmath}
\usepackage[nointegrals]{wasysym}
\usepackage[final]{acl}

\usepackage{times}
\usepackage{latexsym}

\usepackage[T1]{fontenc}

\usepackage[utf8]{inputenc}

\usepackage{microtype}

\usepackage{inconsolata}
\usepackage{multirow}
\usepackage{booktabs}
\usepackage{amssymb}

\usepackage{xcolor}
\usepackage{graphicx}
\newcommand{\cmark}{\CIRCLE}
\newcommand{\pmark}{\LEFTcircle}
\newcommand{\xmark}{\Circle}

\title{StreamMemBench: Streaming Evaluation of Agent Memory for Future-Oriented Assistance}

\author{
  \textbf{Guanming Liu\textsuperscript{1}},
  \textbf{Yuqi Ren\textsuperscript{1}},
  \textbf{Hansu Gu\textsuperscript{2}},
  \textbf{Peng Zhang\textsuperscript{1}}\thanks{Corresponding authors},
\\
  \textbf{Weihang Wang\textsuperscript{1}},
  \textbf{Jiahao Liu\textsuperscript{1}},
  \textbf{Ning Gu\textsuperscript{1}},
  \textbf{Tun Lu\textsuperscript{1}}\footnotemark[1]
\\
\\
  \textsuperscript{1}Fudan University, Shanghai, China \quad
  \textsuperscript{2}Independent Researcher, Seattle, United States
\\
  \texttt{gmliu24@m.fudan.edu.cn};
  \texttt{zhangpeng\_@fudan.edu.cn};
  \texttt{lutun@fudan.edu.cn}
}

\usepackage{eso-pic}
\newcommand{\anthologyFirstPage}{20947}
\AddToShipoutPictureBG{%
  \AtPageLowerLeft{%
    \put(\LenToUnit{0.5\paperwidth+1.1pt},\LenToUnit{50pt}){%
      \makebox(0,0)[b]{\normalsize\thepage}}%
    \ifnum\value{page}=\anthologyFirstPage\relax
      \put(\LenToUnit{0.5\paperwidth},\LenToUnit{34.3pt}){%
        \makebox(0,0)[b]{\footnotesize
          Findings of the Association for Computational Linguistics: EMNLP 2026, pages 20947--20960}}%
      \put(\LenToUnit{0.5\paperwidth},\LenToUnit{23.3pt}){%
        \makebox(0,0)[b]{\footnotesize
          October 24-29, 2026 \copyright 2026 Association for Computational Linguistics}}%
    \fi
  }%
}
\AtBeginDocument{\setcounter{page}{20947}}

\begin{document}
\maketitle
\begin{abstract}
  A central role of personal-agent memory is to turn stored information and prior interactions into future-oriented assistance. In daily use, useful cues come from what the agent observes and how the user interacts with the agent, and the agent must carry them forward from the current request to similar future tasks. Existing memory benchmarks usually test dialogue recall or task improvement in isolation, leaving the trajectory from streaming observations to later assistance largely untested. We introduce StreamMemBench, a streaming benchmark that constructs a two-step task sequence around each evidence anchor from EgoLife egocentric streams. The initial task tests evidence use, while the follow-up task tests whether feedback and interaction experience are reused. Four metrics diagnose evidence recall, initial evidence use, feedback incorporation, and follow-up reuse. Experiments with eight memory systems across two backbones show that current systems often fail to use observed evidence or turn feedback into reliable follow-up behavior, even when evidence is stored or feedback is incorporated locally. StreamMemBench is publicly available at \url{https://github.com/landian60/StreamMemBench}.

\end{abstract}

\section{Introduction}

\begin{figure}[t]
\centering
\includegraphics[width=0.5\textwidth]{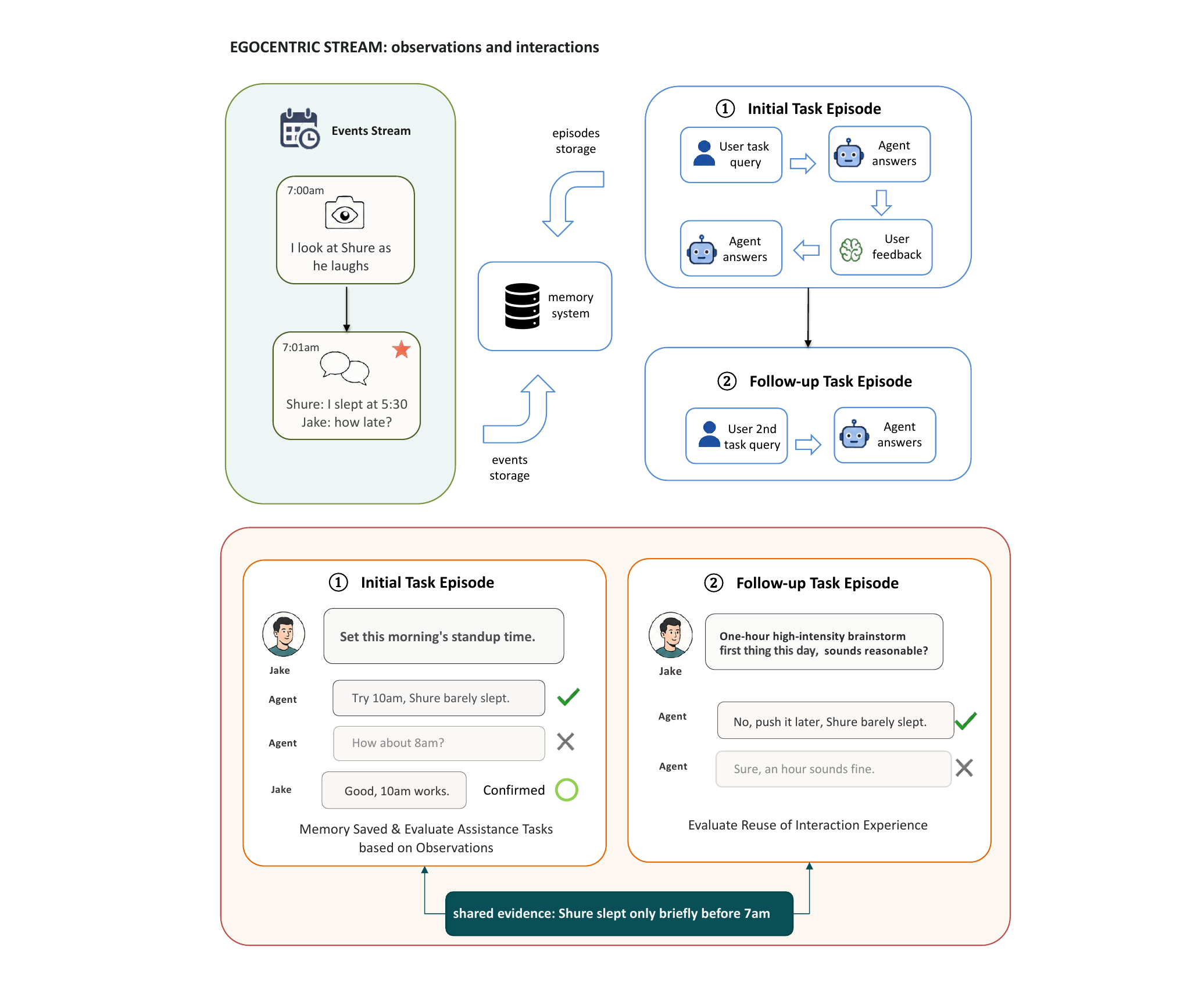}
\caption{A streaming view of personal-agent memory. A memory system identifies and stores user-related information from an egocentric stream to support an initial task. It then stores the resulting interaction, including user feedback, to support future assistance in a related follow-up task.}
\label{fig:motivation}
\end{figure}

Memory is widely recognized as a foundational capability for personal agents built on large language models (LLMs). Beyond retaining user-specific information such as preferences, habits, relationships, and prior interactions~\citep{hu2026evermemos, chhikara2025mem0, kang2025memoryos, li2025memos}, memory mechanism should also enable this information to be applied to future tasks. However, most existing memory agents are unable to accomplish this goal. This limitation is especially important when personal agents operate over continuous streams. To enable future assistance, an agent must identify user-related information from egocentric observations and interactions, store it, use it for the current task, and consolidate the resulting feedback for later similar tasks (As shown in Figure~\ref{fig:motivation}). Even in advanced commercial assistants such as ChatGPT and Gemini, stored information may be available but fail to provide effective assistance when needed. This gap between stored memory and future task behavior suggests that personal-agent memory should be examined at two levels. The first level refers to evidence use, where the agent leverages past observations as contextual evidence to support users' current tasks. The second level refers to experience reuse, where the agent generalizes from interactions involving user feedback to handle similar scenarios. Accordingly, evaluating agent memory requires testing not only what they store, but how effectively stored information supports both present assistance and future tasks.

A growing body of benchmarks evaluates long-term memory, but neither evidence use nor experience reuse is fully evaluated. In these benchmarks, the information an agent must retain is usually presented as dialogue history, profiles, or task instructions. This evades the core streaming challenge of identifying what matters from ongoing observations and interactions and later applying it to a task without being explicitly prompted. Because most datasets use scripted dialogues or synthesized chats~\citep{maharana2024locomo, jiang2025personamem, wu2025longmemeval}, this kind of evidence use goes unevaluated. Similarly, the user feedback that drives experience reuse is usually synthetic and not tied to verifiable observations~\citep{hu2026memoryagentbench, ai2025memorybench, he2026memoryarena}. Without evidence grounding, it is hard to determine whether a correct answer is supported by the observations rather than just sounding reasonable~\citep{wang2025diving}, or whether experience reuse truly reflects actual user involvement. Table~\ref{tab:benchmark-comparison} situates representative benchmarks along five dimensions that capture these distinctions.

\begin{table*}[t]
\centering
\footnotesize
\setlength{\tabcolsep}{4.3pt}
\caption{\textbf{Comparison with representative memory benchmarks.}
  Stream-sourced: real-world observation stream. \quad Evidence Use: extracting and grounding assistance in evidence from observations. \quad User Involvement: user feedback shapes subsequent evaluation. \quad Experience Reuse: consolidating and reusing interaction experience across tasks. \quad Traceability: tracing evidence from its origin in the stream to its use in task behavior.
  \cmark{} = full; \pmark{} = partial; \xmark{} = none.}
\label{tab:benchmark-comparison}
\begin{tabular}{@{}lccccc@{}}
\toprule
Benchmark & Stream-sourced & Evidence Use & User Involvement & Experience Reuse & Traceability \\
\midrule
\multicolumn{6}{@{}l@{}}{\textit{Personal-memory benchmarks}} \\
LoCoMo~\citep{maharana2024locomo} & \xmark{} & \xmark{} & \xmark{} & \xmark{} & \pmark{} \\
PersonaMem~\citep{jiang2025personamem} & \xmark{} & \xmark{} & \xmark{} & \pmark{} & \pmark{} \\
LongMemEval~\citep{wu2025longmemeval} & \xmark{} & \xmark{} & \xmark{} & \xmark{} & \pmark{} \\
EverMemBench~\citep{hu2026evermembench} & \xmark{} & \xmark{} & \xmark{} & \pmark{} & \pmark{} \\
LifeDialBench~\citep{zheng2026lifedialbench} & \pmark{} & \pmark{} & \xmark{} & \xmark{} & \pmark{} \\
\midrule
\multicolumn{6}{@{}l@{}}{\textit{Agent-memory benchmarks}} \\
MemoryAgentBench~\citep{hu2026memoryagentbench} & \xmark{} & \xmark{} & \xmark{} & \pmark{} & \xmark{} \\
MemoryBench~\citep{ai2025memorybench} & \xmark{} & \xmark{} & \pmark{} & \cmark{} & \xmark{} \\
Evo-Memory~\citep{wei2025evomemory} & \xmark{} & \xmark{} & \xmark{} & \cmark{} & \pmark{} \\
MemoryArena~\citep{he2026memoryarena} & \xmark{} & \xmark{} & \xmark{} & \cmark{} & \pmark{} \\
\midrule
\textbf{StreamMemBench} & \textbf{\cmark} & \textbf{\cmark} & \textbf{\cmark} & \textbf{\cmark} & \textbf{\cmark} \\
\bottomrule
\end{tabular}
\end{table*}

We propose \textbf{StreamMemBench}, a streaming benchmark for evaluating whether personal-agent memory supports future-oriented assistance across egocentric observations and interactions. StreamMemBench is built on EgoLife~\citep{egolife2025}, an egocentric dataset that continuously records daily life through wearable devices. From this observation stream we extract user-specific evidence and organize evaluation around \textit{evidence anchors}. Every anchor supports a two-step task sequence. The initial episode gives the agent an assistance request that depends on the anchored evidence, and after the agent responds it receives simulated user feedback. The completed interaction then becomes part of the agent's memory, and a follow-up episode presents a different request grounded in the same anchor to test whether the initial interaction has been consolidated and can be reused. This trajectory is evaluated from both process and performance perspectives. At the process level, we use Fidelity to check whether the anchored evidence is retained before the task, and Feedback Incorporation (FI) to check whether corrective feedback is incorporated within the same interaction. At the performance level, we use Initial Evidence Use (IEU) to measure whether the initial response uses the evidence, and Follow-up Reuse (FUR) to measure whether the follow-up response reuses the evidence or interaction experience. Because every anchor is tied to a specific piece of evidence from the observation stream, each piece of information can be traced from its first appearance to how the agent ultimately uses or misses it, enabling fine-grained diagnosis across all four metrics.


Our experiments show that current memory systems are not yet reliable for future-oriented assistance under streaming evaluation. They often struggle to use evidence from egocentric observations in the initial task and to turn interaction feedback into reusable behavior for the follow-up task. The analysis further shows that retaining evidence alone is insufficient for successful task use, motivating evaluation that traces the path from observed evidence to initial use, feedback incorporation, and follow-up reuse.

Our contributions are:
\begin{itemize}
\item We introduce \textbf{StreamMemBench}, a streaming benchmark for evaluating whether personal-agent memory supports future-oriented assistance across egocentric observations and interactions.

\item We use evidence anchors to test whether agents use the right information in the initial task and reuse it in the follow-up task, and to localize observable failures across evidence retention, initial evidence use, feedback incorporation, and follow-up reuse.

\item We evaluate eight memory systems across three backbones and find that current systems are weak at future-oriented assistance, often failing to use observed evidence or to turn feedback into reliable follow-up behavior.

\end{itemize}

\section{Related Work}

\subsection{Memory Mechanisms for AI Assistants}

Memory mechanisms allow LLM-based personal agents to preserve user preferences, interaction history, task context, and feedback across sessions. Representative systems differ in how they manage stored experience. Retrieval-augmented approaches store past interactions in external repositories and retrieve relevant records during inference, including both dedicated memory stores MemoryBank~\citep{zhong2024memorybank} and general-purpose RAG pipelines~\citep{lewis2020rag}. Memory management approaches treat memory as an actively updated resource where agents decide what to write, update, link, or forget. These systems vary in what they store, spanning fact-level extraction (Mem0~\citep{chhikara2025mem0}, EverMemOS~\citep{hu2026evermemos}), chunk-level linking (A-Mem~\citep{xu2026amem}), and hierarchical consolidation (MemOS~\citep{li2025memos}, MemoryOS~\citep{kang2025memoryos}). Beyond facts and events, procedural memory approaches focus on reusable skills and behavioral patterns accumulated from prior interactions~\citep{zhang2026memskill, liu2026intpro, fang2025memp, yang2026mememind}. Across these designs, a common premise is that storing or managing past experience should improve future assistance. This premise is not guaranteed by storage alone. A system may retain memory without using the right evidence when a later task requires it, or may react to feedback without consolidating the interaction for future reuse.

\subsection{Memory Benchmarks for LLM Agents}

Personal-memory benchmarks evaluate whether systems retain and reason over user histories expressed as dialogues and profiles. For example, LoCoMo and PersonaMem test conversational recall and dynamic user profiling over multi-session interactions~\citep{maharana2024locomo, jiang2025personamem}. Across memory and multimodal settings, benchmarks extend evaluation along context length, participant structure, and diagnostic granularity, as exemplified by LongMemEval, EverMemBench, and MemeBench, respectively~\citep{wu2025longmemeval, hu2026evermembench, wang2026memebench}. Lifelog benchmarks such as EgoLife and LifeDialBench are closer to streaming personal-agent settings because they ground information in realistic egocentric observation streams~\citep{egolife2025, zheng2026lifedialbench}, but they do not test whether observed evidence and interaction feedback jointly shape future assistance. Agent-memory benchmarks evaluate whether memory improves later behavior in interactive or continual settings. MemoryAgentBench tests whether memory improves task performance through incremental interactions and user feedback~\citep{hu2026memoryagentbench}, while MemoryBench and MemoryArena extend this principle to continual learning and interdependent multi-session tasks~\citep{ai2025memorybench, he2026memoryarena}. These benchmarks test behavioral improvement, but their feedback is usually not tied to verifiable observations from the same stream. StreamMemBench differs by anchoring each trajectory in evidence extracted from an egocentric lifelog, tracing whether agents use observational evidence and consolidate interaction feedback into future behavior.

\begin{figure*}[t]
\centering
\includegraphics[width=\textwidth]{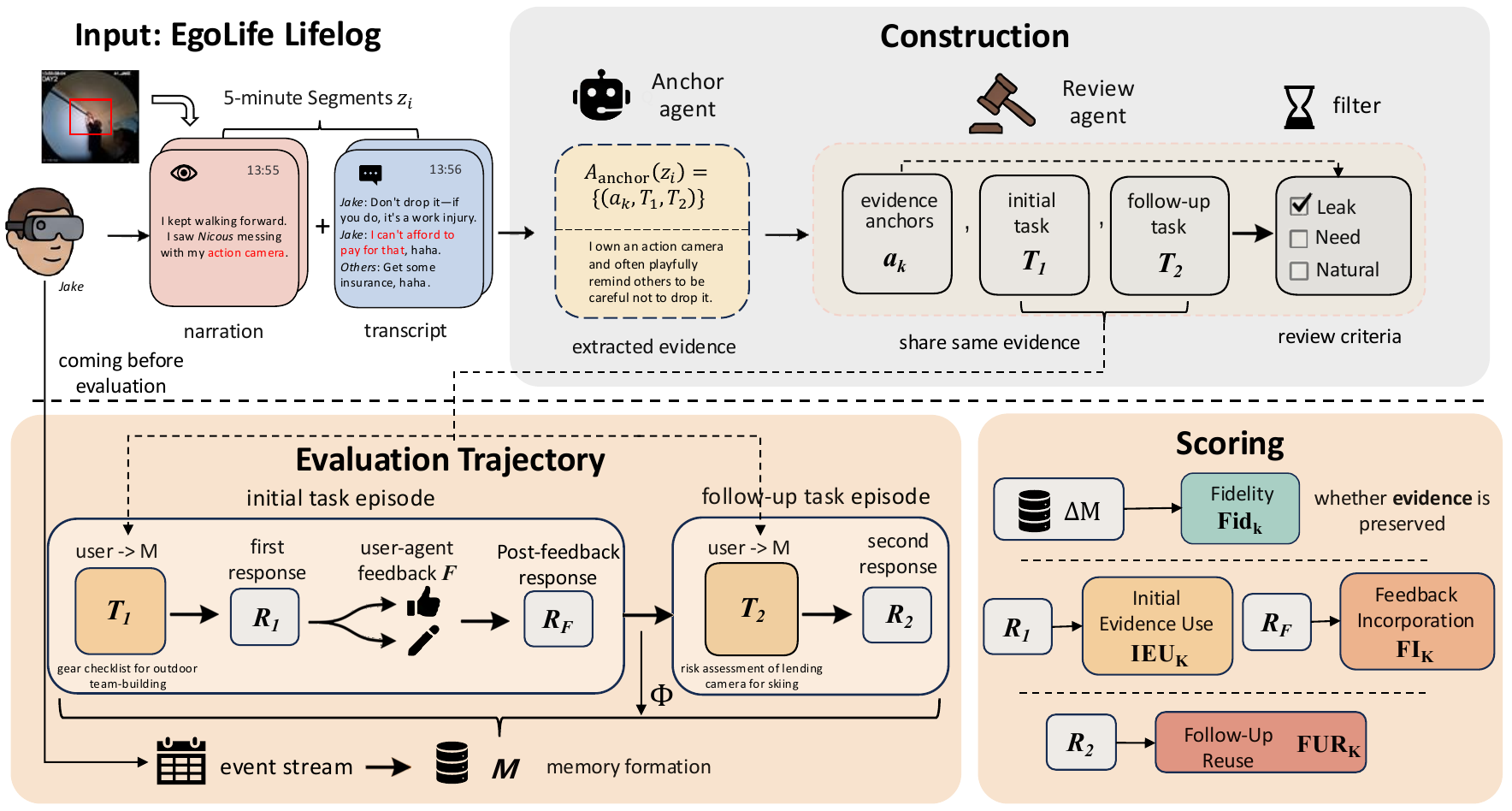}
\caption{\textbf{Benchmark overview.} StreamMemBench first converts EgoLife lifelog segments into evidence anchors with connected initial and follow-up tasks, then evaluates a memory system on a chronological trajectory. The system observes the stream, responds to the initial task, receives feedback, commits the interaction to memory, and answers the follow-up task. The hidden evidence anchor supports four scores that diagnose evidence recall, initial evidence use, feedback incorporation, and follow-up reuse.}
\label{fig:benchmark_overview}
\end{figure*}

\section{Method}
\label{sec:method}

StreamMemBench has two stages, construction and evaluation. The construction stage turns EgoLife lifelog segments into evidence anchors with two-step task sequences. The evaluation stage runs the corresponding task trajectories against a memory system to measure evidence use and experience reuse.

In the construction stage, an evidence anchor agent $A_{\text{anchor}}$ processes each segment and constructs candidate triples consisting of an evidence anchor and two task queries. The anchor records user-specific evidence extracted from the segment and links it to supporting observations. The first query acts as the initial task, and the second the follow-up task. Correct responses to both queries depend on the anchored evidence, which is never explicitly presented. A review agent $A_{\text{review}}$ checks evidence support, consistency with the supporting observations, and query quality. Only candidates that pass this review are retained for evaluation.

In the evaluation stage, the memory system $M$ first ingests the lifelog stream $L$ chronologically. $M$ then receives the initial task and produces a response. A user agent $A_{\text{user}}$ simulates user feedback, either confirming the response or correcting it by supplying the missing evidence. The completed initial episode, containing the task, the response, and the feedback, is committed to memory alongside the lifelog context. $M$ then receives the follow-up task, which tests whether the system has consolidated both the evidence from $L$ and the experience from the initial interaction. An evaluation agent $A_{\text{eval}}$ scores fidelity against atom-level checklists derived from the evidence anchor. Only $M$ is evaluated.

\subsection{Benchmark Construction}

\paragraph{EgoLife segments.}
The lifelog stream $L = (z_1, \ldots, z_n)$ is a chronological sequence of five-minute EgoLife segments. Each segment preserves the order of its observations $x_{i,j}$, each of which is either an egocentric narration or a dialogue transcript. The original stream order is retained so that evidence anchors can point back to their source observations. During evaluation, $M$ receives the segments in chronological order, with observations inside each segment kept in their original stream order.

\paragraph{Evidence anchor construction.}
$A_{\text{anchor}}$ processes the observations in each segment $z_i$ in stream order and constructs candidate triples $(a_k,T_1,T_2)$. The anchor $a_k$ contains one piece of user-specific evidence and the source observations that support it. Evidence is retained only when it captures user-specific information with reusable implications for future assistance, such as a capability, experience, relationship, preference, plan, commitment, or constraint. The two queries are application-oriented requests generated around the same evidence. Correct responses to both queries depend on the anchored evidence, which is never explicitly revealed in the query text. A segment may yield multiple triples when it contains distinct evidence. Observed actions are also retained when they reveal stable or actionable information, whereas isolated scene actions with no reusable implication are excluded. Weakly grounded, over-inferred, or locally inconsistent candidates are filtered by $A_{\text{review}}$.

For each input $z_i$, $A_{\text{anchor}}$ writes a set of evidence anchors and associated task queries.

\[
A_{\text{anchor}}( z_i) = \{(a_k, T_1, T_2)\}.
\]

Each candidate $(a_k,T_1,T_2)$ is checked by a review agent $A_{\text{review}}$ against its source segment for evidence support, subject attribution, consistency with the supporting observations, and three core query constraints.

\[
\begin{aligned}
\operatorname{Leak}(T_i,a_k)&=0,\\
\operatorname{Need}(T_i,a_k)&=1,\\
\operatorname{Natural}(T_i)&=1,
\end{aligned}
\]
for $i \in \{1,2\}$. $\operatorname{Leak}=0$ means the query does not reveal the extracted evidence. $\operatorname{Need}=1$ means an answer that ignores $a_k$ will be incomplete, generic, or wrong. $\operatorname{Natural}=1$ means the query resembles a plausible applied request to a personal agent rather than a factual recall request. The review also checks that the two queries rely on the same anchor while presenting different application scenarios.

\subsection{Evaluation Trajectory and Scoring}
Each evaluation trajectory is scored at four points. Before the task sequence, Fidelity tests whether the target evidence can be recovered. During the task sequence, IEU, FI, and FUR score the initial response, the post-feedback response, and the follow-up response respectively.

For each constructed anchor, StreamMemBench runs one evaluation trajectory against $M$. The trajectory begins with the lifelog segments from the start of $L$ up to the segment that $a_k$ was extracted from, followed immediately by the task sequence, all in a single chronological stream through $M$'s normal interaction interface. Task-side evaluation is black-box. We do not inspect $M$'s internal memory state. Task scores use only $M$'s text responses, while Fidelity is computed separately from the memory delta that $M$ exposes during stream ingestion.

For each anchor, $M$ steps through the following sequence (denoted $\tau_k$):
\[
T_1 \;\rightarrow\; R_1 \;\rightarrow\; F \;\rightarrow\; R_F \;\rightarrow\; \Phi,
\]
\[
\Phi \;\rightarrow\; T_2 \;\rightarrow\; R_2.
\]
$T_1$ and $T_2$ are the initial and follow-up tasks; $R_1$ and $R_2$ are $M$'s responses; $F$ is the user feedback; $R_F$ is the post-feedback response; $\Phi$ commits the completed interaction $(T_1,R_1,F,R_F)$ to $M$ through its own memory-formation interface (session memory, an explicit save call, or a retrieval update).

\paragraph{Fidelity check.}
Fidelity is computed before the task sequence from the memory deltas that $M$ exposes during stream ingestion. After each segment $z_i$, we collect a memory delta $\Delta M_i$, the records newly written or updated by $M$ through its memory interface. For an anchor $a_k$ extracted from $z_i$, $A_\text{eval}$ compares $\Delta M_i$ with the target evidence in $a_k$ and passes only when the same information is present with the correct participant attribution, temporal scope, and factuality:
\[
\text{Fid}_k = \mathbf{1}[A_\text{eval}(\Delta M_i, a_k)=\text{pass}].
\]
The match is semantic rather than verbatim, so paraphrases and sufficiently specific summaries can pass, while missing evidence, contradictions, or wrong attribution fail. This check runs separately from the task trajectory and measures whether the evidence is preserved in the exposed memory delta. It does not show that $M$ can use the evidence in an open-ended task.

\paragraph{User simulator and task scoring.}
Task-side evaluation is handled by $A_\text{user}$. For each task response, $A_\text{user}$ compares the answer with the expected behavior for the query and the evidence in $a_k$. A response passes only when it uses the anchor evidence appropriately; omissions, fabrications, or contradictions fail. After $R_1$, $A_\text{user}$ returns structured output containing natural-language feedback $F$ and a pass/fail task score. This score gives IEU and sets the feedback type $\rho_k \in \{\text{confirmation},\text{correction}\}$. When $\rho_k = \text{correction}$, the feedback supplies the missing or corrected evidence and $A_\text{user}$ scores whether $R_F$ incorporates it as FI. After $T_2$, $A_\text{user}$ scores whether $R_2$ reuses the evidence or corrected experience as FUR. The three task-side scores are
\[
\begin{aligned}
\text{IEU}_k &= \mathbf{1}[A_\text{user}(R_1,T_1,a_k)=\text{pass}],\\
\text{FI}_k   &= \mathbf{1}[A_\text{user}(R_F,F,a_k)=\text{pass}],\\
\text{FUR}_k  &= \mathbf{1}[A_\text{user}(R_2,T_2,a_k)=\text{pass}],
\end{aligned}
\]
where $\text{FI}_k$ is computed only over anchors with $\rho_k = \text{correction}$. The evaluation agent $A_\text{eval}$ is used only for the Fidelity check above.

\section{Experiments}

We evaluate two retrieval baselines and six memory systems on StreamMemBench. We report main results on the four metrics, analyze how performance changes as the stream grows and how systems respond to feedback, validate benchmark construction and scoring against human judgments, categorize failure patterns, and compare storage and computational cost.

\subsection{Experimental Setup}

\paragraph{Data.}
StreamMemBench is built on EgoLife~\citep{egolife2025}, a dataset of 7-day continuous egocentric recordings from 6 participants. In this benchmark, we divide original data into 3347 five-minute segments. Each segment contains roughly 12 timestamped observations, each being either a short description of what the wearer sees and does, or a transcribed utterance from conversations around the wearer. From this stream we extract 7760 evidence anchors, spanning directly stated facts (6712) and contextually inferred knowledge (1048). Two task queries are generated per anchor, producing 15,520 queries that cover task assignment, activity planning, social communication, plan evaluation, gift suggestion, and other natural assistance types. Each evaluation trajectory ingests the stream prefix up to the segment containing the anchor under test.

\paragraph{Evaluated systems.} We evaluate two retrieval baselines and six memory systems that differ in what they store and how they update it. RAG$_{\text{raw}}$ stores the stream as raw segments and retrieves the most relevant ones at query time without modifying them. RAG$_{\text{ext}}$ adds an extraction step where an LLM first pulls structured facts from the stream, and retrieval then operates over those facts instead of the original segments. The six memory systems each actively manage stored content. Mem0~\citep{chhikara2025mem0} stores individual facts and overwrites them when new information conflicts. EverMemOS~\citep{hu2026evermemos} preserves richer structure, keeping event descriptions, personal traits, and situational context as separate layers rather than merging everything into one fact set. A-Mem~\citep{xu2026amem} keeps the stream content largely intact but links related observations together, adding new connections as more segments arrive rather than rewriting. MemOS~\citep{li2025memos} and MemoryOS~\citep{kang2025memoryos} both consolidate information into multiple levels. MemOS refreshes each level on its own schedule by rebuilding it from the level below. MemoryOS sorts content by how recently it appeared, maintaining separate short-, mid-, and long-term stores that refresh at different rates. MemSkill~\citep{zhang2026memskill} differs in kind, storing reusable skills extracted from past interactions rather than facts or observations about the user.

\paragraph{Backbones.} Each memory system is run on \textbf{DeepSeek-V4-Flash (Preview)}, \textbf{Gemini-3-Flash}, and \textbf{Qwen3-14B}. Decoding is deterministic (temperature 0, fixed seed), which keeps $R_1$ reproducible across branches of the same anchor. $A_{\text{anchor}}$, $A_{\text{review}}$, $A_{\text{eval}}$ and $A_{\text{user}}$ use DeepSeek-V4-Pro (Preview).

\paragraph{Streaming evaluation.} Each system receives the segments in time order, up to the segment that contains the anchor. Fidelity is measured by directly extracting the memory delta after each segment and checking whether the anchor's evidence appears in it. Initial evidence use, feedback incorporation, and follow-up reuse are scored by $A_\text{user}$ as described in §3. All $A_\text{user}$ and $A_\text{eval}$ judgments run three times per anchor and are decided by majority vote. Vector-based retrieval paths use the \texttt{BAAI/bge-m3} encoder. We set the shared answer-time return budget to top-$k=10$ while preserving each system's native retrieval mechanism. Appendix~\ref{sec:topk} reports a sensitivity check over $k\in\{3,5,10\}$.

\subsection{Main Results}

Table~\ref{tab:main} reports four pass rates for DeepSeek-V4-Flash (Preview) and Gemini-3-Flash. Table~\ref{tab:qwen} reports the corresponding results for Qwen3-14B.

\begin{table*}[t]
\centering
\small
\setlength{\tabcolsep}{10pt}
\renewcommand{\arraystretch}{1.3}
\caption{\textbf{Main results on StreamMemBench.} Fidelity and FI are process diagnosis metrics scored by $A_\text{eval}$ and $A_\text{user}$ respectively. IEU (Initial Evidence Use) and FUR (Follow-up Reuse) are task performance metrics scored by $A_\text{user}$. $\dagger$: Fidelity inflated by raw-text retention in stored memories.}
\label{tab:main}
\begin{tabular}{lccccccccc}
\toprule
& \multicolumn{4}{c}{DeepSeek-V4-Flash (Preview)} & \multicolumn{4}{c}{Gemini-3-Flash} \\
\cmidrule(lr){2-5} \cmidrule(lr){6-9}
& \multicolumn{2}{c}{Diagnosis} & \multicolumn{2}{c}{Performance} & \multicolumn{2}{c}{Diagnosis} & \multicolumn{2}{c}{Performance} \\
\cmidrule(lr){2-3} \cmidrule(lr){4-5} \cmidrule(lr){6-7} \cmidrule(lr){8-9}
System & Fidelity & FI & IEU & FUR & Fidelity & FI & IEU & FUR \\
\midrule
\multicolumn{9}{@{}l@{}}{\textit{Retrieval baselines}} \\
RAG$_{\text{raw}}$  & 100.0\textsuperscript{$\dagger$} & 76.45 & 27.95 & 60.96 & 100.0\textsuperscript{$\dagger$} & 65.64 & 41.98 & 43.96 \\
RAG$_{\text{ext}}$  & 71.51 & 65.22 & 30.92 & 41.06 & 64.76 & 75.58 & 17.95 & 23.96 \\
\midrule
\multicolumn{9}{@{}l@{}}{\textit{Agentic memory}} \\
Mem0         & 60.68 & 72.33 & 34.95 & 40.96 & 51.61 & 64.93 & 25.94 & 27.95 \\
EverMemOS    & 44.87 & 76.94 & 35.02 & 48.94 & 46.37 & 74.37 & 21.98 & 44.03 \\
A-Mem        & 100.0\textsuperscript{$\dagger$} & 84.63 & 34.92 & 64.98 & 100.0\textsuperscript{$\dagger$} & 79.91 & 30.03 & 56.04 \\
MemOS        & 67.97 & 77.25 & 2.94 & 3.96 & 39.51 & 68.83 & 3.96 & 5.97 \\
MemoryOS     & 100.0\textsuperscript{$\dagger$} & 80.22 & 23.93 & 61.95 & 100.0\textsuperscript{$\dagger$} & 71.97 & 35.94 & 53.96 \\
\midrule
\multicolumn{9}{@{}l@{}}{\textit{Procedural memory}} \\
MemSkill     & 52.58 & 74.07 & 18.94 & 36.04 & 35.31 & 60.53 & 13.96 & 27.95 \\
\bottomrule
\end{tabular}
\end{table*}

The main trends are consistent across the two backbones in Table~\ref{tab:main}. Systems with perfect or high Fidelity can still have much lower IEU and FUR, showing that a Fidelity pass does not guarantee successful task use. A-Mem and MemoryOS remain among the stronger follow-up systems, while MemOS remains low on both IEU and FUR despite relatively high FI. Qwen3-14B results in Table~\ref{tab:qwen} preserve the same separation: RAG$_{\text{raw}}$, A-Mem, and MemoryOS reach perfect Fidelity but substantially lower IEU and FUR, while MemOS remains low on both task-side metrics.

\subsection{Analysis}

\paragraph{Evidence retention vs initial task use.} Table~\ref{tab:main} shows a substantial gap between preserving the target evidence in memory and using it in the initial task. RAG$_{\text{ext}}$ records 71.51\% Fidelity and 30.92\% IEU with DeepSeek-V4-Flash (Preview), and 64.76\% Fidelity and 17.95\% IEU with Gemini-3-Flash. The same pattern appears for Mem0, which records 60.68\% Fidelity and 34.95\% IEU with DeepSeek-V4-Flash (Preview), and 51.61\% Fidelity and 25.94\% IEU with Gemini-3-Flash. A Fidelity pass paired with an IEU failure indicates that the target evidence is present in the exposed pre-task memory delta but is not used successfully in the initial response. This gap may arise from evidence granularity, compression, organization, or difficulty applying the stored evidence to the task.

\begin{figure}[t]
\centering
\includegraphics[width=\columnwidth]{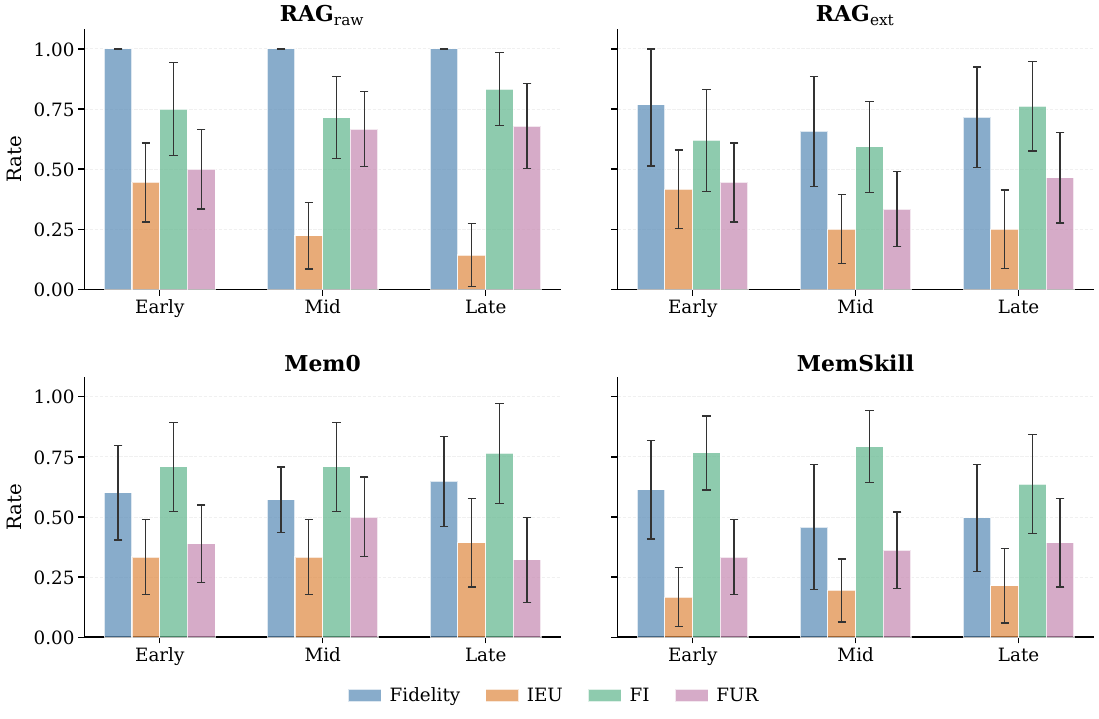}
\caption{\textbf{Capability curves across stream positions.} DeepSeek-V4-Flash (Preview) results are shown. Each panel shows one representative system over early, middle, and late anchors in the chronological stream. The four curves separate evidence availability, initial evidence use, feedback incorporation, and follow-up reuse.}
\label{fig:memory_capability_curves}
\end{figure}

\paragraph{Temporal drift over the stream.} Figure~\ref{fig:memory_capability_curves} shows that the gap between evidence availability and task use changes as the stream grows. RAG$_{\text{raw}}$ keeps Fidelity flat at the top of the plot, but its IEU falls steadily from early to late anchors. This suggests that raw retrieval can keep evidence accessible while becoming less reliable at using it in the initial response. Its FI and FUR move upward after the early split, showing that feedback and later tasks can recover part of the initial use loss. RAG$_{\text{ext}}$ shows a related pattern, with late FI rising again after the middle split. Mem0 is more stable on Fidelity, IEU, and FI, but its FUR weakens late in the stream, suggesting that stable fact storage does not guarantee durable cross turn reuse. MemSkill shows the complementary limitation. Its IEU and FUR improve slightly, while late Fidelity and FI decline, indicating weaker evidence retention and correction incorporation. These trends support the main claim of StreamMemBench. Memory systems should be evaluated by tracing the full path from stored evidence to initial task use, post-feedback response, and later reuse, because a single aggregate score would hide temporal drift across these abilities.

\paragraph{Feedback effect.} The FI and FUR columns of Table~\ref{tab:main} compare immediate correction with later reuse. FI is computed only on correction cases, where $R_1$ fails and the user feedback supplies the missing evidence. It measures whether the system can revise its behavior within the same turn. FUR is computed after the post-feedback response and interaction commit, and measures whether the evidence or corrected experience supports a related follow-up task. Across all eight systems, FUR exceeds IEU by an average of 18.66 points with DeepSeek-V4-Flash (Preview) and 11.51 points with Gemini-3-Flash. A paired ablation on 20 anchors per system separates the two sources of this difference: without the $T_1$ interaction commit, FUR is 27.5\%, only 4.4 points above IEU; committing the same interaction before the identical $T_2$ query raises FUR to 40.6\%, accounting for the remaining 13.1 points (Appendix~\ref{sec:commit-ablation}). Low IEU and high FI on correction cases show that explicit feedback often repairs the immediate response, while the commit ablation shows that retaining the intervening interaction improves FUR, especially after correction.

\begin{figure}[t]
\centering
\includegraphics[width=0.95\columnwidth]{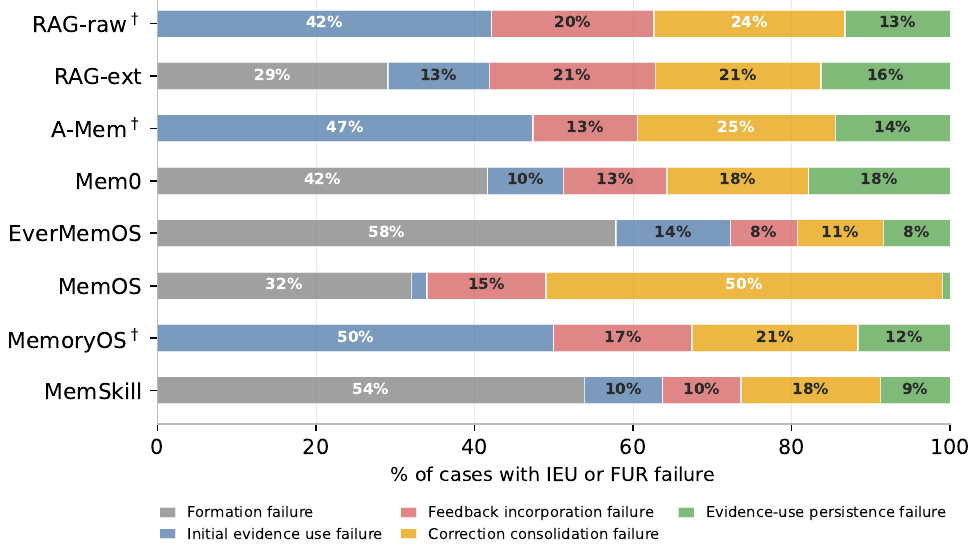}
\caption{\textbf{Lifecycle failure modes for failed evidence use or reuse.} Bars are normalized over evaluation trajectories with IEU = 0 or FUR = 0. Each trajectory is assigned one failure-mode label based on its Fidelity, IEU, FI, and FUR outcomes. Systems marked $^\dagger$ have no Formation failure cases among the displayed trajectories.}
\label{fig:failure_modes}
\end{figure}

\paragraph{Lifecycle failure modes.} We classify each evaluation trajectory with IEU = 0 or FUR = 0 into one of five mutually exclusive failure modes that locate where the evidence-to-behavior path breaks down. These labels summarize observable score patterns along the evaluation trajectory. If the target evidence is not recoverable by the Fidelity check, the trajectory is labeled Formation failure. Otherwise, its label is determined by the IEU, FI, and FUR outcomes. Initial evidence use failure means that the initial response misses the evidence, feedback repairs the behavior, and the follow-up succeeds. Feedback incorporation failure means that the initial response fails and the system also fails to use the corrective feedback within the same turn. Correction consolidation failure denotes cases where corrective feedback is incorporated immediately but the follow-up still fails. Evidence-use persistence failure means that the initial task uses the evidence correctly but the follow-up task fails to reuse it.

Figure~\ref{fig:failure_modes} shows that the breakdown point differs by memory mechanism. For systems marked with $^\dagger$, none of the displayed trajectories is a Formation failure; their plotted cases are therefore distinguished by the subsequent IEU, FI, and FUR outcomes. This matches the main table pattern in which high Fidelity does not guarantee high IEU or FUR. Extraction-based and fact-based systems show larger Formation failure shares, indicating that a larger proportion of the plotted trajectories also fail the pre-task Fidelity check. MemOS is dominated by Correction consolidation failure, which is consistent with its high FI but very low FUR in Table~\ref{tab:main}. A manual inspection of representative MemOS failures suggests that many missed cases are not pure write failures. The target facts can appear in saved memories, but the retrieved context for the task does not surface them, leaving the answer stage without the needed evidence. Once the feedback states the missing fact, the same generator can revise the response, a pattern consistent with the higher FI. Overall, the lifecycle view supports the paper's main diagnostic claim. StreamMemBench does not only measure whether a system stores evidence, but locates where the observable path from evidence retention to initial use, post-feedback response, and follow-up reuse breaks down. Appendix~\ref{sec:workstation-case} illustrates these failure patterns with one workstation-setup case by comparing the content exposed through each system's memory interface and its responses at $T_1$, after feedback, and at $T_2$.

\subsection{Human Validation}

We audit both benchmark construction and metric scoring against human judgments. For construction, two annotators independently review 210 items, sampled as five items from each of the 42 participant-days. They assess whether each evidence anchor is supported by the source observations, whether the two task specifications are natural, non-leaking, evidence-dependent, and distinct in scenario, and whether the simulated feedback is plausible. For scoring, we sample 30 complete trajectories from each memory system and compare the Fidelity, IEU, FI, and FUR judgments with human annotations. Table~\ref{tab:human-audit} reports construction pass rates and scoring agreement together with Cohen's $\kappa$.

\begin{table}[t]
\centering
\footnotesize
\setlength{\tabcolsep}{4pt}
\caption{\textbf{Human validation of benchmark construction and scoring.} Pass/agreement is the construction pass rate for the first three rows and agreement with human annotations for the four scoring rows.}
\label{tab:human-audit}
\begin{tabular}{lcc}
\toprule
Audit target & Pass/agreement (\%) & Cohen's $\kappa$ \\
\midrule
Evidence anchor $a_k$ & 85.2 & 0.319 \\
$T_1/T_2$ task specification & 90.7 & 0.350 \\
Feedback $F$ & 93.3 & 0.388 \\
Fidelity scoring & 96.1 & 0.699 \\
IEU scoring & 94.4 & 0.691 \\
FI scoring & 91.7 & 0.646 \\
FUR scoring & 92.8 & 0.608 \\
\bottomrule
\end{tabular}
\end{table}

All seven audit criteria show agreement beyond chance ($\kappa>0$), alongside construction pass rates above 85\% and scoring agreement above 90\%.

\subsection{Storage, Latency, and Token Cost}

Streaming evaluation also exposes system overhead during memory updates and task response. Table~\ref{tab:cost} reports per-segment costs under DeepSeek-V4-Flash (Preview).

\begin{table}[t]
\centering
\small
\setlength{\tabcolsep}{6pt}
\renewcommand{\arraystretch}{1.15}
\caption{\textbf{Storage, latency, and token cost with DeepSeek-V4-Flash (Preview).} Storage is the memory content stored after processing one segment, in bytes. Latency is the time to ingest one segment and answer $T_1$, in seconds. Token cost counts prompt and completion tokens per segment, in thousands.}
\label{tab:cost}
\begin{tabular}{lccc}
\toprule
System & Storage (B) & Latency (s) & Tokens (k) \\
\midrule
RAG$_{\text{raw}}$  & 8044.46          & 5.52              & 9.88 \\
RAG$_{\text{ext}}$  & 1789.11          & 3.34              & 5.34 \\
A-Mem           & 7829.79          & 6.68              & 38.57 \\
Mem0            & 569.36           & 2.52              & 15.15 \\
EverMemOS       & 502.46           & 4.99              & 20.07 \\
MemOS           & 657.71           & 4.02              & 7.77 \\
MemoryOS        & 8286.46          & 5.44              & 38.48 \\
MemSkill        & 699.68           & 3.11              & 8.90 \\
\bottomrule
\end{tabular}
\end{table}

The table shows that storage, latency, and token cost do not move together. Compact memory representations such as EverMemOS, Mem0, MemOS, and MemSkill keep storage below 1 KB per segment, while raw or heavily linked state leads to much larger memory footprints. Latency remains in a narrow range for most systems, with A-Mem the main high-latency outlier. Token cost varies more sharply, especially for A-Mem and MemoryOS. These results suggest that streaming memory evaluation should report multiple efficiency dimensions, since a system can be compact in storage but expensive in tokens, or lightweight in tokens but weaker in task behavior.

\section{Conclusions}

We introduced StreamMemBench to evaluate personal-agent memory as a capability for future-oriented assistance. Rather than treating memory as stored information alone, StreamMemBench tests whether observations and interactions change how an agent helps the user later. Our experiments show that current memory systems remain limited under this view, even when they appear capable of retaining information or responding to feedback in the moment. These results point to a broader evaluation principle: memory should be judged by whether it supports useful future behavior.

\section{Limitations}

The benchmark evaluates a selected set of representative memory systems and retrieval baselines rather than the full space of commercial and research memory agents. Its evidence comes from the current EgoLife setting, which does not establish generalizability across broader linguistic and cultural contexts. Future work can extend the benchmark to longer and more diverse egocentric streams and broader system families.

Potential risks come from the personal nature of egocentric observations and memory evaluation. Although StreamMemBench is intended for research on safer and more reliable personal agents, benchmarks of this kind may encourage systems to store or infer sensitive user information. Deployment should therefore include consent, data minimization, access control, and mechanisms for users to inspect, correct, and delete memories. Evaluation results should not be interpreted as a license to deploy persistent memory in settings where privacy and user control are not well defined.

\section*{Acknowledgments}

This work is supported by the Major Project of the National Social Science Fund of China (NSSFC) under Grant No. 25\&ZD260. Peng Zhang and Tun Lu are with the College of Computer Science and Artificial Intelligence, Fudan University. Tun Lu is also affiliated with the MOE Laboratory for National Development and Intelligent Governance and the Shanghai Key Laboratory of Data Science, Fudan University.

\bibliography{custom}

\newpage
\clearpage    
\appendix
\section{License Information}

StreamMemBench is intended for non-commercial academic research. Commercial use of the dataset, model, or any derivatives is prohibited.

\section{Intended Use Compliance}

StreamMemBench is constructed from the publicly available EgoLife dataset under its terms for non-commercial academic research.

\section{Additional Backbone Results}

Table~\ref{tab:qwen} reports the full Qwen3-14B results under the same evaluation protocol as the two primary backbones.

\begin{table}[h]
\centering
\footnotesize
\setlength{\tabcolsep}{5pt}
\caption{\textbf{Results with Qwen3-14B.} All values are pass rates. Systems marked with $\dagger$ retain source text directly in memory.}
\label{tab:qwen}
\begin{tabular}{lcccc}
\toprule
System & Fidelity & FI & IEU & FUR \\
\midrule
RAG$_{\text{raw}}$ & 100.00\textsuperscript{$\dagger$} & 63.04 & 16.99 & 22.02 \\
RAG$_{\text{ext}}$ & 51.73 & 59.35 & 14.78 & 19.72 \\
Mem0 & 49.09 & 57.78 & 17.63 & 20.77 \\
EverMemOS & 34.95 & 56.52 & 15.42 & 22.28 \\
A-Mem & 100.00\textsuperscript{$\dagger$} & 73.91 & 15.78 & 27.32 \\
MemOS & 33.64 & 67.35 & 2.12 & 5.43 \\
MemoryOS & 100.00\textsuperscript{$\dagger$} & 71.74 & 17.17 & 22.18 \\
MemSkill & 38.13 & 59.57 & 13.63 & 20.13 \\
\bottomrule
\end{tabular}
\end{table}

\section{Interaction-Commit Ablation}
\label{sec:commit-ablation}

We run a paired ablation on 160 trajectories, with 20 anchors per system. Each trajectory branches from the same memory checkpoint before the completed $T_1$ interaction is committed. In \textsc{No-commit}, the identical $T_2$ query is evaluated without saving the preceding $(T_1,R_1,F,R_F)$ interaction, and the branch remains read-only. In \textsc{Full}, the same interaction is committed through the system's memory-formation interface ($\Phi$) before evaluating $T_2$. The two branches preserve the same observational memory and differ only in whether $\Phi$ commits the intervening interaction before $T_2$.

\begin{table}[t]
\centering
\scriptsize
\setlength{\tabcolsep}{6pt}
\caption{\textbf{Interaction-commit ablation on a paired subset.} IEU and FUR are pass rates (\%); Commit gain is Full FUR minus No-commit FUR in percentage points. The average is the macro mean over systems, computed before rounding.}
\label{tab:commit-ablation}
\begin{tabular}{lcccc}
\toprule
System & IEU & No-commit FUR & Full FUR & Commit gain \\
\midrule
RAG$_{\text{raw}}$ & 50.0 & 35.0 & 55.0 & +20.0 \\
RAG$_{\text{ext}}$ & 20.0 & 30.0 & 40.0 & +10.0 \\
Mem0 & 25.0 & 30.0 & 40.0 & +10.0 \\
EverMemOS & 10.0 & 30.0 & 30.0 & 0.0 \\
A-Mem & 35.0 & 40.0 & 45.0 & +5.0 \\
MemOS & 5.0 & 5.0 & 10.0 & +5.0 \\
MemoryOS & 30.0 & 30.0 & 60.0 & +30.0 \\
MemSkill & 10.0 & 20.0 & 45.0 & +25.0 \\
\midrule
\textbf{Average} & \textbf{23.1} & \textbf{27.5} & \textbf{40.6} & \textbf{+13.1} \\
\bottomrule
\end{tabular}
\end{table}

On the No-commit branch, FUR exceeds IEU by 4.4 points; 24 trajectories fail on $T_1$ but pass on $T_2$, while 17 show the opposite outcome. Committing the interaction adds 13.1 points to FUR under the same $T_2$ query, producing the 17.5-point difference between IEU and Full FUR.

\begin{table}[t]
\centering
\scriptsize
\setlength{\tabcolsep}{3pt}
\caption{\textbf{Commit gain by $T_1$ feedback type.} No-commit and Full report FUR pass rates (\%); gains are computed from unrounded counts.}
\label{tab:commit-feedback}
\begin{tabular}{lcccc}
\toprule
Feedback after $T_1$ & $n$ & No-commit & Full & Gain \\
\midrule
Correction & 123 & 19.5 & 35.0 & +15.4 \\
Confirmation & 37 & 54.1 & 59.5 & +5.4 \\
\bottomrule
\end{tabular}
\end{table}

On this subset, Commit gain is larger after correction than after confirmation (15.4 vs.\ 5.4 points), consistent with the intervening interaction contributing more when the initial response requires revision.

\section{Retrieval-Budget Sensitivity}
\label{sec:topk}

We vary the shared answer-time return budget over $k\in\{3,5,10\}$ on a fixed 100-sample subset using DeepSeek-V4-Flash (Preview), with all other settings unchanged. Table~\ref{tab:topk-sensitivity} reports IEU, FUR, and query-answer time for all eight systems. On the eight-system macro average, increasing $k$ from 3 to 10 raises IEU from 23.8\% to 29.3\% and FUR from 43.3\% to 49.6\%, while mean query-answer time increases from 4.156 to 4.455~s. Across systems, $k=10$ gives the highest or tied-highest IEU and the highest FUR.

\begin{table*}[t]
\centering
\scriptsize
\caption{\textbf{Retrieval-budget sensitivity on a 100-sample subset with DeepSeek-V4-Flash (Preview).} IEU and FUR are pass rates (\%); query-answer time is measured in seconds. The average is the macro mean over systems.}
\label{tab:topk-sensitivity}
\begin{tabular}{lccccccccc}
\toprule
& \multicolumn{3}{c}{IEU} & \multicolumn{3}{c}{FUR} & \multicolumn{3}{c}{Query-answer time} \\
\cmidrule(lr){2-4} \cmidrule(lr){5-7} \cmidrule(lr){8-10}
System & $k=3$ & $k=5$ & $k=10$ & $k=3$ & $k=5$ & $k=10$ & $k=3$ & $k=5$ & $k=10$ \\
\midrule
RAG$_{\text{raw}}$ & 24.0 & 28.0 & 28.0 & 54.0 & 58.0 & 60.0 & 3.740 & 3.810 & 3.996 \\
RAG$_{\text{ext}}$ & 26.0 & 32.0 & 32.0 & 54.0 & 58.0 & 60.0 & 1.950 & 1.910 & 2.180 \\
Mem0 & 28.0 & 32.0 & 34.0 & 36.0 & 38.0 & 41.0 & 10.580 & 10.930 & 11.310 \\
EverMemOS & 28.0 & 32.0 & 36.0 & 54.0 & 56.0 & 59.0 & 2.650 & 2.750 & 2.720 \\
A-Mem & 24.0 & 32.0 & 34.0 & 42.0 & 50.0 & 51.0 & 4.890 & 4.970 & 5.063 \\
MemOS & 18.0 & 22.0 & 22.0 & 24.0 & 26.0 & 27.0 & 2.680 & 2.740 & 2.990 \\
MemoryOS & 24.0 & 26.0 & 27.0 & 52.0 & 59.0 & 62.0 & 4.470 & 4.550 & 4.710 \\
MemSkill & 18.0 & 21.0 & 21.0 & 30.0 & 36.0 & 37.0 & 2.290 & 2.380 & 2.670 \\
\midrule
\textbf{Average} & \textbf{23.8} & \textbf{28.1} & \textbf{29.3} & \textbf{43.3} & \textbf{47.6} & \textbf{49.6} & \textbf{4.156} & \textbf{4.255} & \textbf{4.455} \\
\bottomrule
\end{tabular}
\end{table*}

\section{Cross-Evaluator Agreement}

We re-score the DeepSeek-V4-Flash (Preview) trajectories with Claude Sonnet 5. Across Fidelity, FI, IEU, and FUR, the average absolute difference between evaluators ranges from 2.15 to 5.32 points, and the system-level rank correlation ranges from 0.67 to 0.95 (Table~\ref{tab:evaluator-robustness}).

\begin{table*}[t]
\centering
\scriptsize
\caption{\textbf{Cross-evaluator comparison on DeepSeek-V4-Flash (Preview) trajectories.} The two judgment sets use DeepSeek-V4-Pro (Preview) and Claude Sonnet 5.}
\label{tab:evaluator-robustness}
\begin{tabular}{lcccccccc}
\toprule
& \multicolumn{4}{c}{DeepSeek-V4-Pro (Preview)} & \multicolumn{4}{c}{Claude Sonnet 5} \\
\cmidrule(lr){2-5} \cmidrule(lr){6-9}
System & Fidelity & FI & IEU & FUR & Fidelity & FI & IEU & FUR \\
\midrule
RAG$_{\text{raw}}$ & 100.00 & 76.45 & 27.95 & 60.96 & 100.00 & 82.85 & 30.12 & 61.21 \\
RAG$_{\text{ext}}$ & 71.51 & 65.22 & 30.92 & 41.06 & 72.44 & 73.68 & 26.33 & 42.07 \\
Mem0 & 60.68 & 72.33 & 34.95 & 40.96 & 62.44 & 64.28 & 27.58 & 44.34 \\
EverMemOS & 44.87 & 76.94 & 35.02 & 48.94 & 47.26 & 80.79 & 31.05 & 53.92 \\
A-Mem & 100.00 & 84.63 & 34.92 & 64.98 & 100.00 & 78.12 & 36.67 & 55.41 \\
MemOS & 67.97 & 77.25 & 2.94 & 3.96 & 62.20 & 82.35 & 2.29 & 5.88 \\
MemoryOS & 100.00 & 80.22 & 23.93 & 61.95 & 100.00 & 82.86 & 28.72 & 60.73 \\
MemSkill & 52.58 & 74.07 & 18.94 & 36.04 & 46.25 & 75.61 & 17.88 & 29.32 \\
\midrule
Avg. absolute difference & -- & -- & -- & -- & 2.15 & 5.32 & 3.29 & 3.63 \\
Rank correlation & -- & -- & -- & -- & 0.95 & 0.67 & 0.69 & 0.88 \\
\bottomrule
\end{tabular}
\end{table*}

\section{Cross-System Workstation Case}
\label{sec:workstation-case}

We trace one workstation-setup evaluation case involving Jake through all eight systems. Both tasks depend on three linked requirements: each participant's glasses must be connected to the designated workstation, data must be transferred to an external drive every three hours, and both drive cables must be checked. The content exposed through each system's memory interface takes different forms. RAG$_{\text{raw}}$ and A-Mem retain raw first-person records; EverMemOS stores a compact episode summary; Mem0 stores six fact memories; MemoryOS stores layered memories; MemOS stores three structured summaries; MemSkill stores nine short memory items; and RAG$_{\text{ext}}$ stores five extracted facts.

The systems also diverge across the task trajectory. MemOS is the only system that passes $T_1$ directly. After feedback, several systems revise their responses; RAG$_{\text{raw}}$ and EverMemOS still omit required constraints. At $T_2$, only A-Mem and MemoryOS pass by applying the participant-to-workstation mapping.

\begin{figure}
    \centering
    \includegraphics[width=0.5\linewidth]{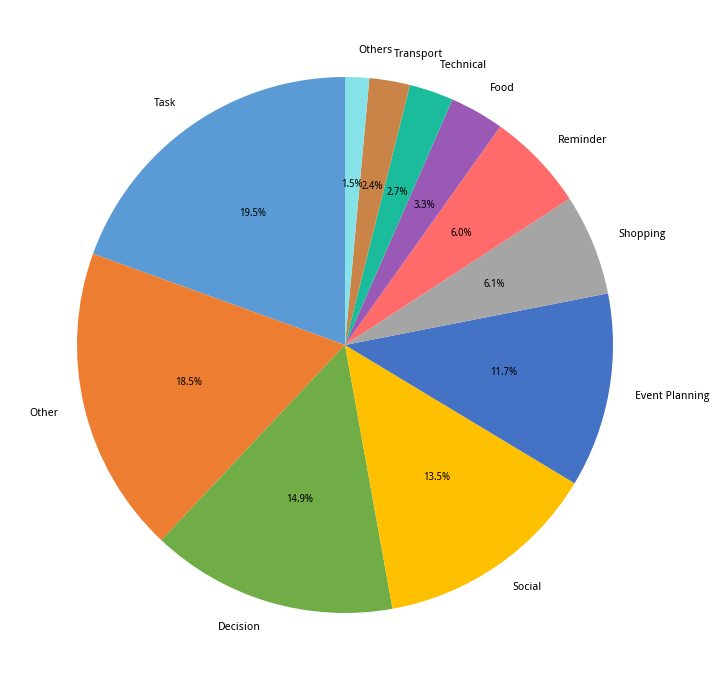}
    \caption{Distribution of task scenarios in the StreamMemBench dataset.}
    \label{fig:distribute}
\end{figure}

\section{Dataset Example Cases}
\begin{figure}
    \centering
    \includegraphics[width=1\linewidth]{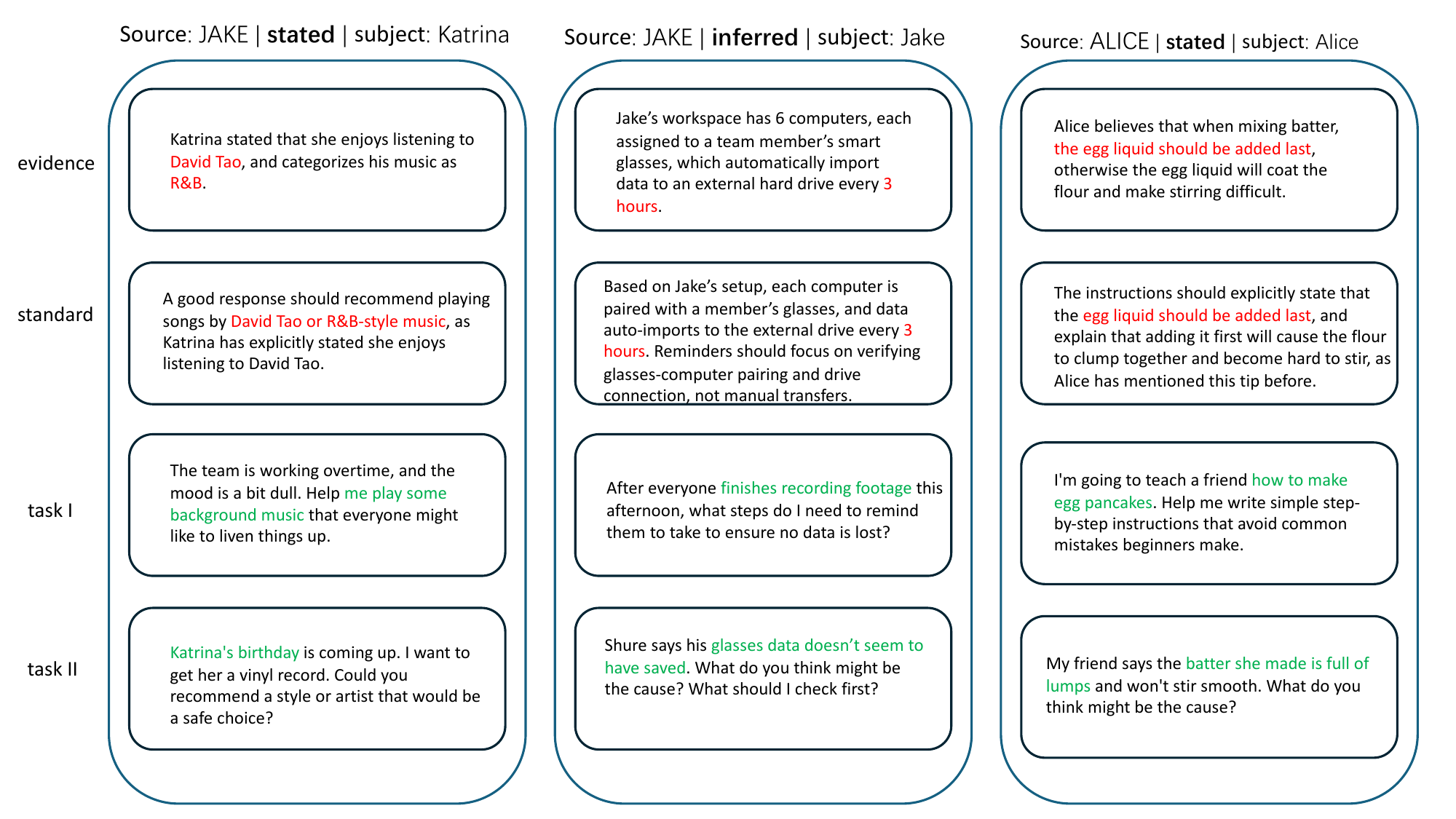}
    \caption{Representative cases with source evidence, expected behavior, and two evidence-dependent downstream tasks, grouped by evidence type (stated/inferred) and subject (self/other).}
    \label{fig:visualize}
\end{figure}

Figure~\ref{fig:visualize} presents representative cases grouped by evidence type (stated/inferred) and subject (self/other). Each case contains the source snippet, expected behavior, and two evidence-dependent downstream tasks; red highlights mark source evidence and green highlights mark task intent.

\section{Dataset Composition}
Figure~\ref{fig:distribute} groups the tasks into ten scenario types. Directly stated evidence accounts for 86.5\% of the evidence anchors, while contextually inferred evidence accounts for 13.5\%. By subject, 47.4\% of queries concern the user's own preferences, routines, or needs, and 52.6\% concern other people in the user's social context.

\section{Prompt Overview}

\subsection{Construction: Evidence anchor and task generation}
Given a lifelog segment, the construction agent extracts source-grounded Evidence Anchors that are useful in future interactions. Each anchor records an \texttt{evidence\_statement}, its subject and evidence type, supporting observations and source span, and two applied tasks with a \texttt{user\_request} and \texttt{expected\_behavior}. The prompt favors transferable information with clear counterfactual value, excludes transient or unsupported content, and requires distinct future scenarios that do not reveal the target evidence.

\subsection{Construction: Evidence revision review}
The review agent checks each round-one anchor for source support, participant attribution, stated-versus-inferred evidence strength, temporal stability, overgeneralization, task naturalness and diversity, and whether the expected behavior stays within the evidence. It returns concise, anchor-specific guidance for the second construction round without creating new anchors or making keep-or-drop decisions.

\subsection{Evaluation: Memory fidelity audit}
The Fidelity prompt compares each ground-truth \texttt{evidence\_statement} with text exposed by the memory system. It accepts semantically equivalent paraphrases or faithful summaries, and rejects absent, contradictory, wrongly attributed, or unsupported extrapolations. The judge returns an item-level \texttt{present} decision with a brief reason.

\subsection{Evaluation: User feedback simulation}
The user simulator compares each task response with the Evidence Anchor and \texttt{expected\_behavior}. It returns \texttt{affirm} when the answer uses the evidence and satisfies the task, or \texttt{revise} with short, natural feedback that supplies the missing or corrected information. The feedback avoids benchmark terminology and reads like a user's direct response to the assistant.

\end{document}